\documentclass[sigconf, nonacm]{acmart}
\usepackage{amsmath}
\usepackage{xcolor}
\usepackage{graphicx}
\usepackage{subcaption}
\usepackage{algorithm}
\usepackage{algorithmic}
\begin{document}
\title{Universal Embeddings of Tabular Data}

%%
%% The "author" command and its associated commands are used to define the authors and their affiliations.
\author{Astrid Franz}
\affiliation{%
  \institution{CONTACT Software}
  \city{Bremen}
  \country{Germany}
}
\email{Astrid.Franz@contact-software.com}

\author{Frederik Hoppe}
\affiliation{%
  \institution{CONTACT Software}
  \city{Bremen}
  \country{Germany}
}
\email{Frederik.Hoppe@contact-software.com}

\author{Marianne Michaelis}
\affiliation{%
  \institution{CONTACT Software}
  \city{Bremen}
  \country{Germany}
}
\email{Marianne.Michaelis@contact-software.com}

\author{Udo G\"obel}
\affiliation{%
  \institution{CONTACT Software}
  \city{Bremen}
  \country{Germany}
}
\email{Udo.Goebel@contact-software.com}

\begin{abstract}
Tabular data in relational databases represents a significant portion of industrial data. Hence, analyzing and interpreting tabular data is of utmost importance. Application tasks on tabular data are manifold and are often not specified when setting up an industrial database. To address this, we present a novel framework for generating universal, i.e., task-independent embeddings of tabular data for performing downstream tasks without predefined targets. Our method transforms tabular data into a graph structure, leverages Graph Auto-Encoders to create entity embeddings, which are subsequently aggregated to obtain embeddings for each table row, i.e., each data sample. This two-step approach has the advantage that unseen samples, consisting of similar entities, can be embedded without additional training. Downstream tasks such as regression, classification or outlier detection, can then be performed by applying a distance-based similarity measure in the embedding space. Experiments on real-world datasets demonstrate that our method achieves superior performance compared to existing universal tabular data embedding techniques.
\end{abstract}

\maketitle

\section{Introduction}\label{sec:intro}
Over the past few decades, companies have been collecting a substantial amount of data, resulting in vast repositories of information. The recent development of numerous analytical tools for analyzing and interpreting data has transformed these data collections into invaluable resources for various applications. However, a significant portion of the data is stored in relational databases, often dispersed across multiple tables, which presents challenges for efficient analysis and interpretation. Despite advancements in data-driven methods utilizing large language models, processing tabular data in an automated way remains challenging.  Due to the lack of common standards and diverging preferences regarding the information to be stored, the challenges persist even when common database software is employed. As a result, data within database tables are typically highly heterogeneous. This heterogeneity is further rooted in the diversity of features stored in tables: the simplest being numerical features, which possess an intrinsic order, while categorical features consist of a specific, unordered set of possible values. Additionally, other modalities, such as text or images, usually appear in tables, adding to their complexity.

For database software providers, serving clients focused on gaining valuable data insights, the main difficulty arises from not having a well-defined use case established at the beginning. This includes the absence of a specified task as well as a related target. Application tasks on tabular data are manifold, ranging from classification and regression to data integration and outlier detection. When setting up an industrial database, the kind of task and even the quantity of interest, i.e. the target, often remain unknown. Facing this challenge, our goal is to create universal embeddings for tabular data serving as a foundation for a wide range of application tasks without predefined targets. The majority of tasks are approached by measuring the similarity between table entries. Therefore, the entries are embedded into a vector space and similarity is evaluated by applying a distance measure in this embedding space. Consequently, various downstream tasks like classification, regression or retrieval can be performed through nearest neighbor search in the embedding space. Thus, our contributions are twofold:
\begin{enumerate}
	\item We develop a framework for creating universal, i.e., target- and task-independent embeddings for the table entities. Subsequently, we aggregate the embedded entities, creating embeddings that represent individual table rows.  
	\item We evaluate the performance of our method with real-world datasets, demonstrating and analyzing superior performance compared to state-of-the-art universal tabular data tools.
\end{enumerate}
The approach of constructing both entity and row embeddings provides a flexible and cost-effective way to embed unseen table rows, consisting of similar entities. This paves the way for a wide range of application tasks, even when the target is not known a priori. Our method is capable of simultaneously processing both categorical and numerical data, effectively accounting for intra- and inter-feature dependencies across a wide range of tables.

\section{Related Work}\label{sec:related work}

In the following, we provide a non-exhaustive overview of crucial directions of research, partly aligned with this work. The surveys \cite{singh2023embeddingstabulardatasurvey, surveyBorisov, jiang2025representationlearningtabulardata} provide a more comprehensive overview of representation learning for tabular data. In our study, we differ from the conventional approach seen in many related works, where embeddings are primarily generated by training a model on a predefined supervised learning task. This methodology optimizes the embeddings for specific tasks, but their effectiveness often diminishes significantly in alternative contexts. In contrast, our approach constructs embeddings that are independent of any particular task, enhancing general applicability. However, this broader applicability comes at a cost: a marginal decrease in performance for specific tasks, compared to embeddings that have been tailored and optimized specifically for those tasks.

\textbf{Tree-based models.}
In contrast to other domains, e.g., computer vision (CV), natural language processing (NLP) and signal processing (SP), tabular data processing and interpretation is still dominated by classical machine learning methods, in particular, gradient boosting decision trees. Gradient Boosting methods such as XGBoost~\cite{XGBoost}, 
LightGBM~\cite{LightGBM} and CatBoost~\cite{CatBoost} have been established as primary tools to analyze 
structured data.

\textbf{Attention-based models.}
Recent research efforts aim to exploit the transformer architecture~\cite{Attentionisallyouneed} in tabular data. Several new methods adapted and extended this architecture in order to analyze tabular data~\cite{Tabnet2021,Tabtransformer,saint,ctBert,TURL,yin-etal-2020-tabert}. However, despite the transformer's tremendous success in NLP, the performance of attention-based methods for tabular data remains below expectations and seems to be tailored to specific tasks and datasets~\cite{whyTreeBetterDL,DLnotallyouneed}. 

\textbf{Tabular Foundation Models.}
Recent tabular foundation models (TFMs) such as TabPFN \cite{hollmann2025tabpfn}, CARTE \cite{CARTE}, TARTE \cite{TARTE} and TabICL \cite{TabICL} pre-train massive transformer backbones and then reuse the model weights to perform in-context learning or light-weight fine-tuning. In doing so they encapsulate both (i) representation learning and (ii) task-specific inference inside a single, resource-intensive network that must be called at inference time. Our objective is not to deliver an end-to-end predictor that is invoked (or fine-tuned or prompted) for each new task, but to learn a task-agnostic vector representation for each entity and each row that can be cached once in a vector store and used for arbitrary downstream tasks. Thus, our goal is to decouple representation learning and task-specific inference. Whereas TFMs provide universal predictors, we would like to provide universal embeddings. The two approaches are complementary and address different engineering questions.

\textbf{MLP-based models.}
In contrast to transformers, there are approaches applying simpler network architectures such as multilayer perceptrons (MLP).
In~\cite{gorishniy2025tabm}, the authors presented TabM, an MLP-based architecture using parameter-efficient ensembling~\cite{Wen2020BatchEnsemble}. This architecture demonstrates an improved performance and higher efficiency compared to attention-based techniques. More recently, \cite{chernov2025ggmoevsmlp} developed a mixture-of-expert approach combined with a gating function that outperforms standard MLP models. However, all these models are trained for a specific predefined task.

\textbf{Universal Embeddings.}
The paper \cite{embdi2020} proposes universal embeddings for tabular data by representing the table as a graph, followed by the construction of sentences via random walks. Then, these sentences are processed by methods from the NLP domain \cite{word2vec,glove} to obtain embeddings for entities, rows and columns of the table. Although it is mainly applied to data integration scenarios, the embeddings do not depend on a downstream task and may be used in a wide range of real-world applications. This algorithm, called EmbDI (Embedding for Data Integration), is the main method we compare with, as it aligns with the goal of this work.

\textbf{Graph Auto-Encoder.}
The seminal paper \cite{kipf2016variationalgraphautoencoders} introduces a graph auto-encoder (GAE) for link prediction on several popular citation network datasets. The encoder model learns meaningful latent representations for each node, allowing for embedding construction and their further processing. Based on this work, a series of subsequent approaches~\cite{kipf2017semisupervised,GATE,GAEFS} has developed more refined architectures. We refer to~\cite{surveyGNN} for a comprehensive overview of graph neural networks including GAEs. The work \cite{VILLAIZANVALLELADO2024106180} leverages graph neural networks and interaction networks to construct contextual embeddings in an encoder-decoder framework. However, the decoder aims to solve a supervised task, resulting in task-dependent embeddings. 

\section{Universal Embedding Construction}\label{sec:embedding workflow}

\begin{figure*}
\includegraphics[width=\textwidth]{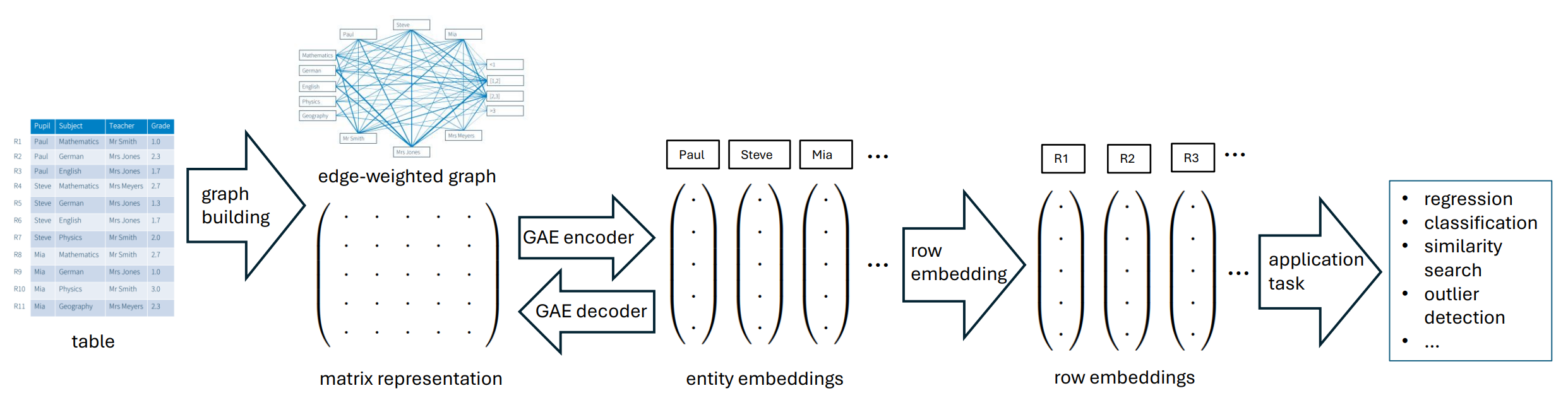}
\caption{Visualization of the proposed algorithm.} 
\label{fig:flowchart}
\end{figure*}

We propose a method for constructing universal embeddings, i.e., embeddings that represent intrinsic table patterns, but do not depend on a predefined target or learning task. This universality ensures their applicability for a wide range of downstream tasks such as regression, classification, similarity search, and outlier detection. Building upon the established strengths of EmbDI \cite{embdi2020} and GAEs \cite{kipf2016variationalgraphautoencoders}, we introduce a novel framework that effectively combines weighted graphs with autoencoding principles, enabling more robust and semantically meaningful representations of tabular data. Algorithm~\ref{alg:embedding_framework} summarizes our methodology for building universal embeddings. Moreover, Figure~\ref{fig:flowchart} illustrates this procedure visually. Initially, we construct a graph where the nodes represent the entities of the table, and this graph is represented as a matrix containing probabilities. Using this matrix, a graph autoencoder seeks to reconstruct it, resulting in embeddings for the entities that capture the inherent structure of the tabular data. By aggregating these entity embeddings, we obtain row embeddings that can be utilized for downstream tasks.

In the following sections, we provide a detailed description of the individual steps in this process.

\begin{algorithm}
\caption{Universal Embedding Construction}
\begin{algorithmic}[1]\label{alg:embedding_framework}
\STATE\textbf{Input:} \\
$\circ$ table entries $(e_{ij})$, 
    $(i,j)\in\{1,\hdots, n_\text{rows}\}\times \{1,\hdots, n_\text{col}\}$\\
$\circ$	embedding dimension $d\in\mathbb{N}$
\STATE Preprocessing: assign numerical values to bins (Section~\ref{subsec:tab_to_graph})\\
    $\rightarrow$ obtain $n$ unique entities (categorical entries and numerical bins) 
\STATE Build edge-weighted graph with $n$ nodes (Sections~\ref{subsec:tab_to_graph} 
    and~\ref{subsec:reduce_graph})
\STATE Derive matrix representation $T\in\mathbb{R}^{n\times n}$ (Section~\ref{subsec:matrix})
\STATE Use regression-based GAE architecture to reconstruct $T$. Encoder outputs the entity embeddings $u_1,\hdots, u_n$ for each node (Section~\ref{subsec:gae})
\STATE\textbf{For} each table row $i = 1,\hdots, n_\text{rows}$:\\
    \hspace*{0.5cm}Calculate row embedding $v_i$ as mean of the corresponding\\
    \hspace*{0.5cm}entity embeddings (Section~\ref{subsec:row_emb}) 
\STATE\textbf{Output:} \\
$\circ$ entity embeddings $u_1,\hdots, u_n \in\mathbb{R}^d$ to embed unseen rows\\
$\circ$ row embeddings $v_1,\hdots,v_{n_\text{rows}}\in\mathbb{R}^d$ for downstream tasks 
\end{algorithmic}
\end{algorithm}

\subsection{From Table to Graph}\label{subsec:tab_to_graph}

\begin{figure}
\includegraphics[height=7.8cm]{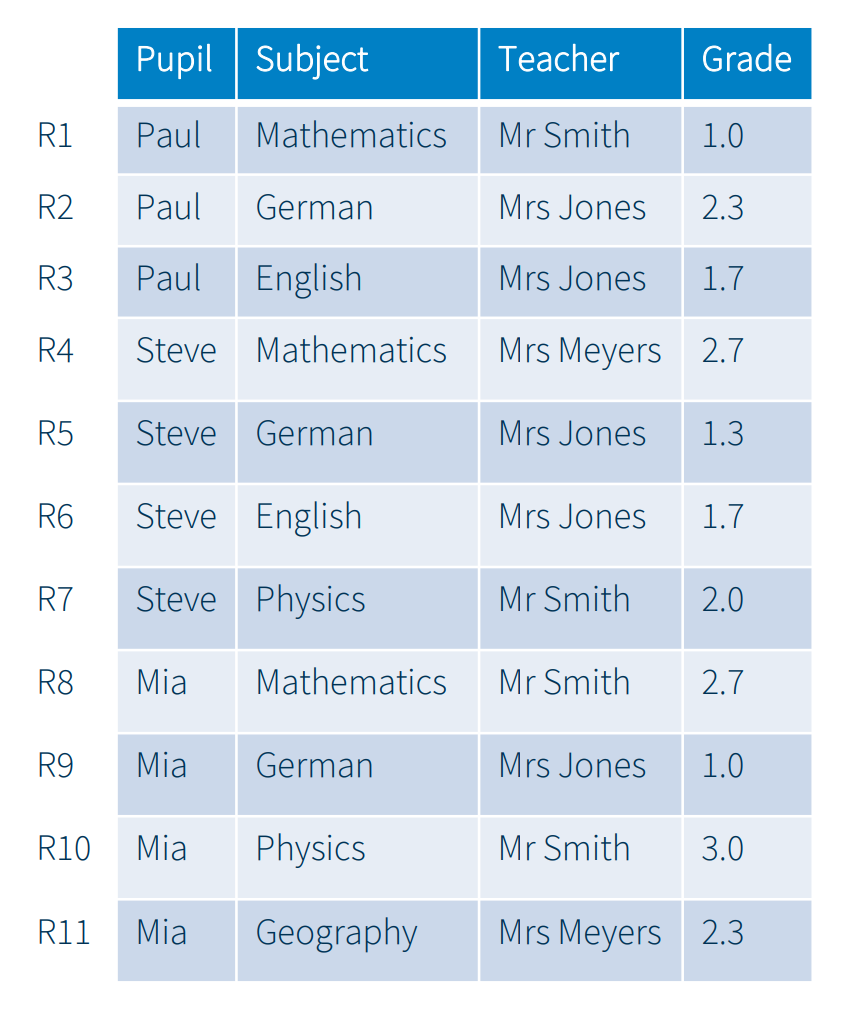}
\includegraphics[height=7.8cm]{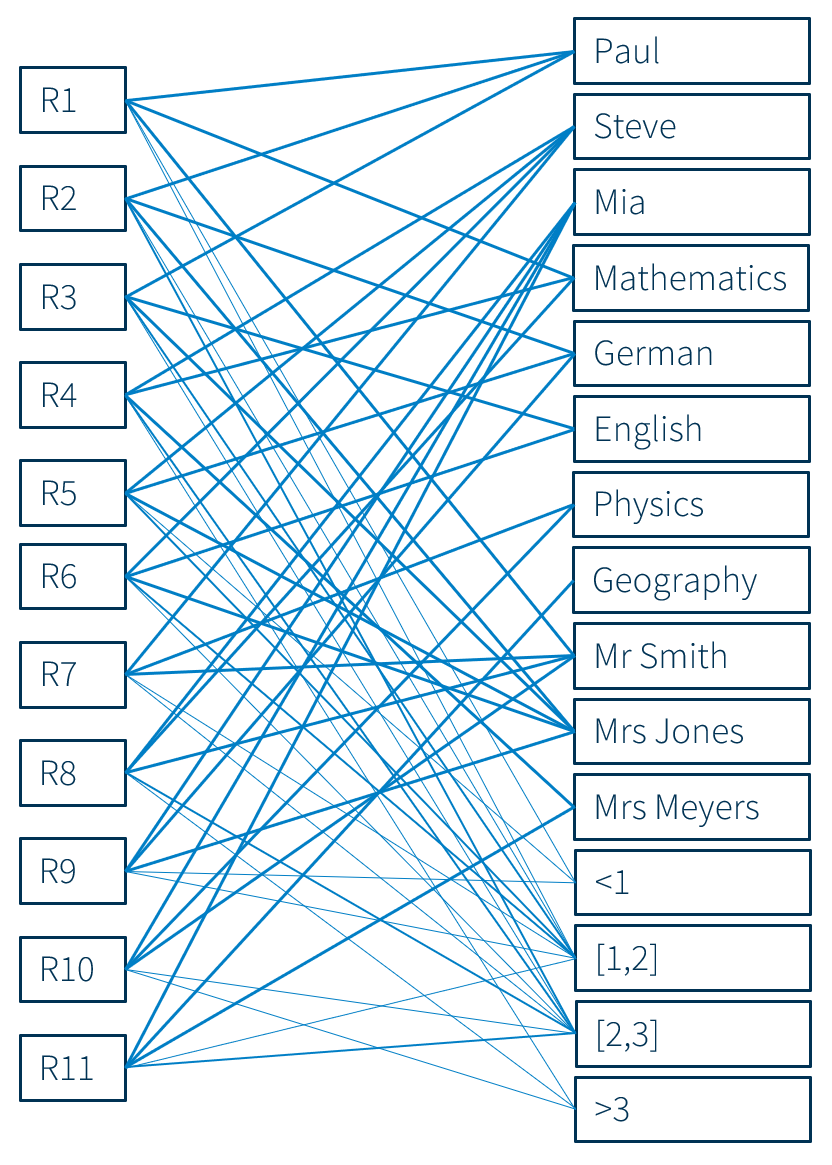}
\caption{A graph is built from a table by nodes representing the row indices and the unique entities. 
Row nodes and entity nodes are connected, if the entities occur in the corresponding row.} 
\label{fig:table_graph}
\end{figure}

In this section, we describe the construction of a graph from a given table. In the following, we use the term \emph{entities} to refer to the unique entries in each categorical column of the table combined with a set of bins arising from the numerical columns. The bins are constructed, such that every bin contains approximately the same number of entries from the corresponding column. The first step, extracting the entities from the table and building a preliminary graph, aligns with the EmbDI algorithm proposed in \cite{embdi2020}. However, we neglect the attribute nodes, as the EmbDI algorithm focuses on data integration as the primary goal, whereas ours is the construction of universal embeddings. Thus, we create a bipartite graph, where one partition consists of all row nodes and the second partition comprises all entity nodes belonging to the categorical and numerical features contained in the table. Figure~\ref{fig:table_graph} illustrates this initial step on a simplified example table. For categorical features, every unique entry in this column represents a node of the graph. Numerical features are summarized in bins, ranging from the minimal to the maximal value, complemented with two additional bins exceeding this range, i.e., one bin for values smaller than the minimal value and another for values larger than the maximal value.

\begin{figure}
\includegraphics[width=\columnwidth]{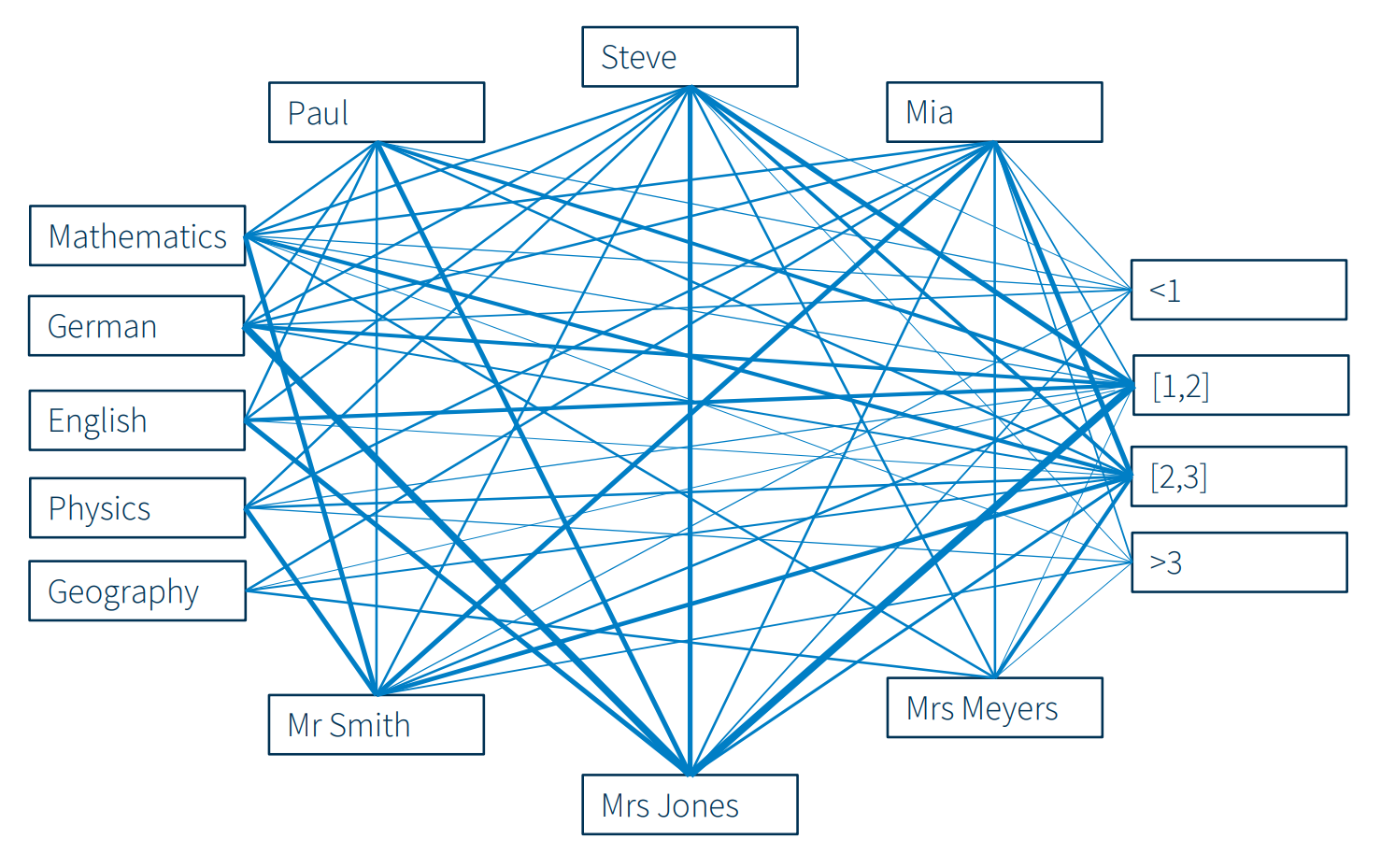}
\caption{Removing the row nodes and connecting entity nodes that are linked via a row node yields a reduced 
graph with weighted edges.}
\label{fig:graph_without_rowindex}
\end{figure}

The edges are built by connecting row nodes and entity nodes with respect to their occurrence in the corresponding table row. For example, if a categorical entity appears in row $i$ of the table, then the corresponding nodes are connected by an edge, which is assigned the weight $1$. For numerical entities, this procedure is more sophisticated. In order to maintain the intrinsic order of numerical values and model their relationship, we consider the exact position of a numerical value $x\in\mathbb{R}$ inside the corresponding bin interval $[b_i,b_{i+1}]\subset \mathbb{R}$. If the value is closer to the lower interval border $b_i$, then the corresponding row node is additionally connected to the node, representing the next lower bin $[b_{i-1}, b_i]$. Conversely, if the value is closer to the upper interval border $b_{i+1}$, then the corresponding row node receives an extra connection to the next higher bin node. The edge weights linking row nodes with the corresponding numerical bin nodes follow the probability function
\begin{equation}
p(x)=\left\{
   \begin{array}{l}
   \displaystyle 0.5 +\frac{x-b_i}{b_{i+1}-b_i} \text{ for } x\leq\frac{b_i+b_{i+1}}{2} \\
   \displaystyle 0.5 + \frac{b_{i+1}-x}{b_{i+1}-b_i} \text{ for } x>\frac{b_i+b_{i+1}}{2}
   \end{array}
   \right..
   \label{eq:bin_prob}
\end{equation}
The edge connecting the row node with the next lower or upper bin node is assigned the weight $1-p(x)$. Thus, a value exactly in the center of a bin, i.e., $x=\frac{b_i+b_{i+1}}{2}$ belongs exclusively to the bin $[b_i, b_{i+1}]$, resulting in an edge weight of 1. The weight 
is increasingly reduced the larger the distance of $x$ to the bin center is. This is illustrated in Figure~\ref{fig:table_graph}, where some edges to the bin nodes (the lower four nodes on the right-hand side) are slightly thinner than the other edges. 
With this proceeding, we capture the information inherent from the intrinsic order of numerical values. Furthermore, since for every numerical value $x$ not only its belonging to a certain bin, but also the exact position inside the bin is captured, no information is lost, and the particular selection of bins is not critical.
Note that the bins are created for every numerical feature separately. This includes that intervals from two different columns which may coincide are treated as two distinct ones. 
Describing tables as graphs in this way has an important advantage: Nodes in a graph are a set without any order. A table with reshuffled rows or columns would result in the same graph as the original table. Hence Properties 1 and 2 of \cite{OBSERVATORY}, describing requirements for tabular embeddings, are intrinsically satisfied.

\subsection{Reducing the Graph}\label{subsec:reduce_graph}

Considering the example in Figure~\ref{fig:table_graph}, the number of row nodes on the left-hand side and the number of entity nodes on the right-hand side of the graph are nearly equal. However, in industrial real-world scenarios, tables may contain thousands or even millions of rows, whereas the number of unique entities is typically much smaller. 

Hence, in contrast to the EmbDI algorithm, we reduce the graph by removing the row nodes. To compensate for this loss of information, the entity nodes are directly linked if they were connected via a row node in the original graph. The edge weight between two nodes is updated with the product of the two former edge weights, connecting the corresponding two nodes with the removed row node. As a consequence, a connection between two entity nodes, where one node represents a numerical bin and the other a categorical entity, is assigned with the weight $p(x)$, where $x$ is the numerical value and $p$ the probability function defined in \eqref{eq:bin_prob}. In case two entities appear simultaneously in multiple rows, the corresponding edge weights are summed. The resulting reduced graph is shown in Figure~\ref{fig:graph_without_rowindex}, where the thickness of an edge illustrates its weight.

This procedure of removing the row nodes significantly reduces the total number of nodes and hence, the computational effort, while preserving the initial table structure. Due to the edge weights, the dynamics of random walks on this reduced graph remain the same as on the original graph. When following the EmbDI approach and building sentences, the major difference is that sentences built on the reduced graph will contain only every other word compared to sentences built on the original graph.

\subsection{Matrix Representation}\label{subsec:matrix}

\begin{figure*}
\includegraphics[width=0.85\textwidth]{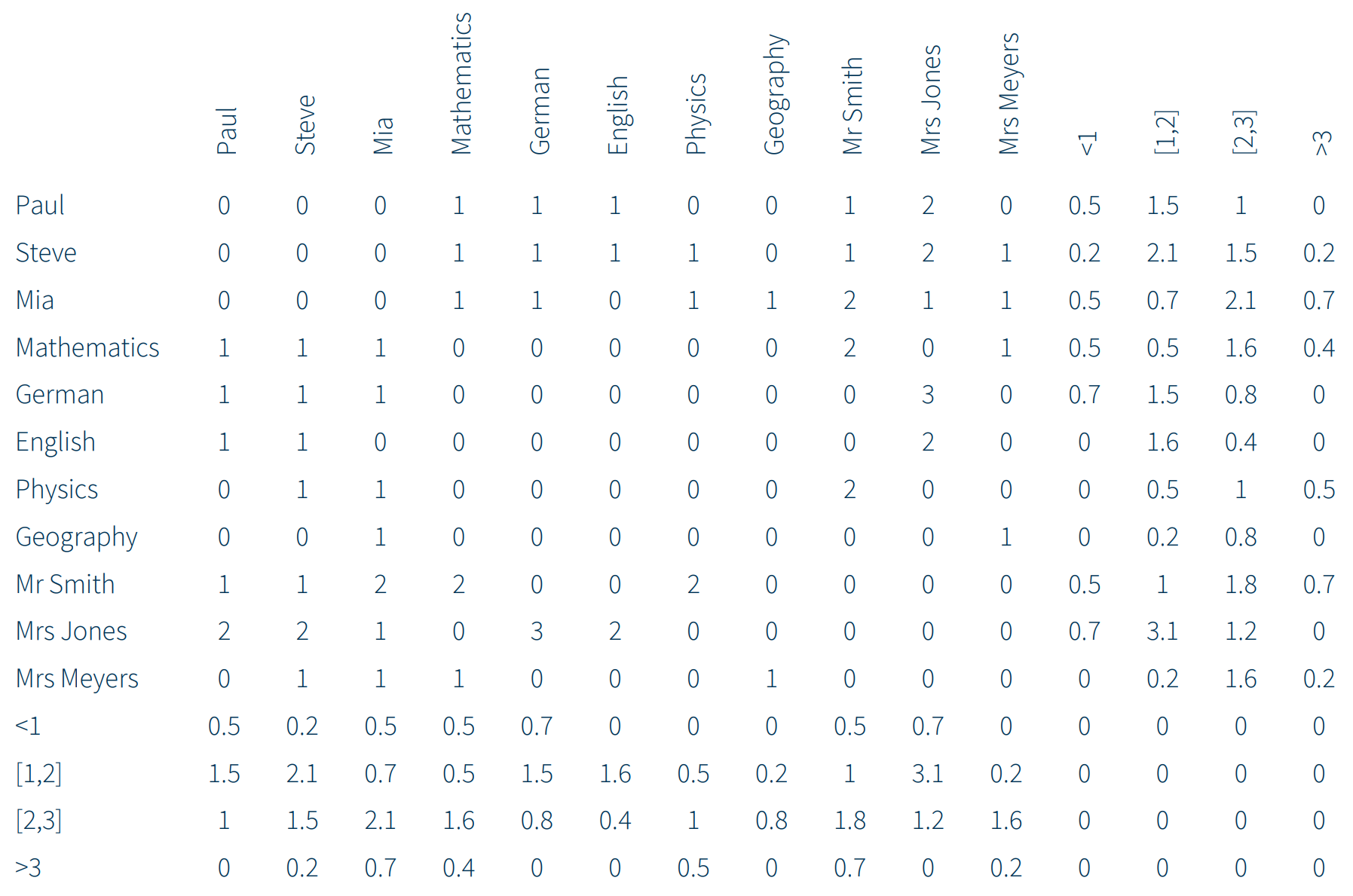}
\caption{Adjacency matrix with absolute edge weights.}
\label{fig:adjacency}
\end{figure*}

A graph structure can be mathematically described using an adjacency matrix, which is a symmetric matrix where the number of rows equals the number of nodes in the graph. Each row and column of the adjacency matrix corresponds uniquely to a graph node. In an unweighted graph, an entry in the adjacency matrix is $1$ if the corresponding nodes are connected and $0$ otherwise. In the case of an edge-weighted graph, like the one considered here, the entry reflects the weight of the edge connecting the corresponding two nodes or $0$, if the nodes are not connected. Figure~\ref{fig:graph_without_rowindex} depicts the reduced graph, and its corresponding adjacency matrix is presented in Figure~\ref{fig:adjacency}.

\begin{figure*}
\includegraphics[width=0.85\textwidth]{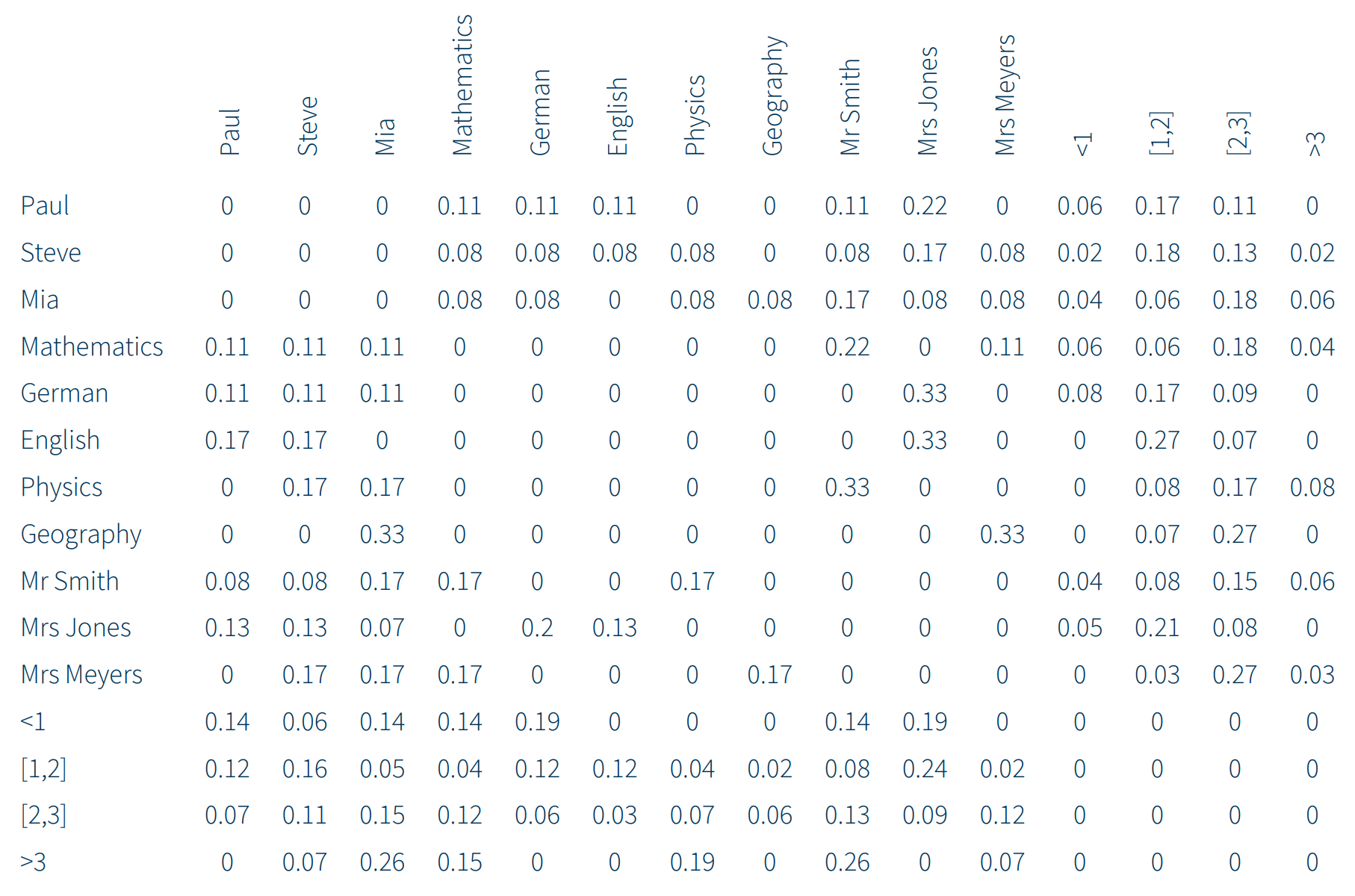}
\caption{Transition matrix containing transition probabilities.}
\label{fig:transition}
\end{figure*}

Through row-wise normalization, the adjacency matrix is transferred into the transition matrix, which is useful for performing random walks on the graph. In the transition matrix, the entry in row $i$ and column $j$ represents the probability of transitioning from node $i$ to node $j$. Figure ~\ref{fig:transition} shows the transition matrix of the adjacency matrix in Figure~\ref{fig:adjacency}.

\subsection{Entity Embeddings}\label{subsec:gae}

In the sequel, for creating embeddings of the entities, represented by $n$ graph nodes, we describe a modified GAE approach, tailored to an edge-weighted graph. The original GAE idea was introduced in \cite{kipf2016variationalgraphautoencoders}. We emphasize that our framework is not limited to this original GAE approach, but applies to more sophisticated GAE architectures in a straightforward way. However, for clear presentation, illustrating the main idea, we consider the original GAE. Moreover, a lot of sequential approaches do not significantly outperform the original approach while consisting of a more complex architecture \cite{GATE}. From the industrial point of view, guaranteeing maintenance plays a major role and favors simpler architectures with similar performance.

The original GAE network reconstructs the binary adjacency matrix of an unweighted graph. We denote the adjacency matrix by $A\in\{0,1\}^{n\times n}$, where $n$ denotes the number of nodes, and an optional node feature matrix by $X\in\mathbb{R}^{n\times d}$, where $d$ is the predefined embedding dimension. The embedding matrix 
$Z\in\mathbb{R}^{n \times d}$, consisting of $n$ $d$-dimensional embeddings, is set as
\begin{equation}
	Z = \operatorname{GCN}(X, A),
\end{equation}
where $\operatorname{GCN}$ is a graph convolutional network \cite{kipf2017semisupervised}. The decoder is defined as the inner product $\hat{A} = \sigma(ZZ^T)$, with the sigmoid function $\sigma$. The key adaption to exploit the information provided by a matrix representation of an edge-weighted graph
is the reconstruction loss. Instead of 
recovering a binary adjacency matrix we aim to retrieve a weighted matrix $T$, i.e.,
\begin{equation}\label{eq:loss}
	\mathcal{L}_{gae} = (1-\alpha) \Vert \hat{T} - T\Vert_2 + \alpha \Vert \hat{T} - T\Vert_1,
\end{equation}
where $\alpha\in (0,1)$ balances the different norms. This refined reconstruction scheme allows for learning 
a representation of the graph that not only accounts for the existence of an edge but also for its weight.

\subsection{Row Embeddings}\label{subsec:row_emb}

Using the GAE described in the previous subsection, we obtain an embedding for each entity in the original table. We denote these entity embeddings by $u_1,\hdots, u_n$. To address application tasks related to the table, we construct embeddings for each row, i.e. for each sample. This is achieved by averaging the entity embeddings of all neighbors of each row node in the original graph (see Section~\ref{subsec:tab_to_graph}), weighted by the corresponding edge weights. When considering the original table, this means computing the embedding of a certain row by averaging the embeddings of all entities occurring in this row. For a numerical value, an average of two bin nodes is taken, weighted with the probability in Equation~(\ref{eq:bin_prob}).

Thus, our method transforms the row tabular data into structured vector spaces which possess an intrinsic order. More precisely, for each sample in the table we get a $d$-dimensional row embedding $v_1,\hdots, v_{n_\text{rows}}$, representing the corresponding row in the vector space $\mathbb{R}^d$. Due to the embedding procedure, the row embeddings of similar rows have a smaller distance to each other than row embeddings of different ones. 

The two-step approach of computing entity embeddings first and then averaging them to obtain row embeddings offers a key advantage: It generalizes naturally to new rows without requiring retraining, as long as they contain previously known entities.

We use a simple mean approach to construct our final embeddings, as it has several advantages over alternative methods. While concatenation would preserve all original information, it leads to higher-dimensional representations that increase computational complexity and storage requirements. Learning-based fusion methods require task-specific training data and optimization objectives, which would inherently bias the resulting embeddings towards specific downstream tasks. This contradicts our goal of creating universal embeddings. Furthermore, concatenation or fusion requires special treatment of missing data. Our mean approach is not influenced by missing data, since it averages over all entities corresponding to a certain row.

\subsection{Evaluating the Embedding Quality}\label{subsec:quality}

The main motivation behind our procedure is its general applicability and independence of the target as well as the downstream task. Due to this independence, there is no natural measure to assess the quality of the embeddings. Nevertheless, for a concrete downstream task, predictions can be constructed based on the row embeddings. The quality of these predictions can serve as an indirect evaluation measure 
for the underlying embeddings.

Therefore, for the two downstream tasks regression and classification we describe exemplary the construction of a prediction. The numerical evaluation with respect to the regression and classification performance will be conducted in Section~\ref{sec:exps}. 

We assume $v_1,\hdots,v_{n_\text{train}}$ to be the embeddings of training rows from the original table. With the same procedure as described in Section~\ref{subsec:row_emb}, we calculate the row embeddings of previously unseen test rows $v_{n_\text{train}+1},\hdots, v_{n_\text{train}+n_\text{test}}$. Furthermore, we assume an additional target table column, indexed with $t=n_{\text{columns}}+1$, which we predict via regression or classification. For each test row embedding $v_l$, we perform a $k$-NN-search to obtain the $k$ nearest neighbors from the training embeddings with respect to the Euclidean distance (or any other suited distance measure) and store the corresponding indices in $K$. The corresponding entities of the target column $t$ are denoted by $e_{it}$, $i\in K$. We merge these $k$ predictions to a single one by weighting them according to the distance to the test embedding, i.e.,
\begin{equation}\label{eq:pred}
	\widehat{e}_{lt} = \sum\limits_{i\in K} \frac{e_{it}}{\Vert v_i - v_l\Vert_2}.
\end{equation}
For regression, the deviation of $\widehat{e}_{lt}$ from $e_{lt}$ for all test indices $l$ 
can be assessed by a variety of quality measures, for instance
\begin{itemize}
\item root mean squared error: 
\begin{equation}
	\text{RMSE}=\sqrt{\frac{1}{n_\text{test}}\sum(e_{lt}-\widehat{e}_{lt})^2},
\end{equation}
\item root mean squared percentage error:
\begin{equation}
	\text{RMSPE}=\sqrt{\frac{1}{n_\text{test}}\sum\left(\frac{e_{lt}-\widehat{e}_{lt}}{e_{lt}}\right)^2}.
\end{equation}
\end{itemize}
The performance of a regression method is the better the smaller these measures are.

For classification on the other hand, Equation~(\ref{eq:pred}) yields a value between 0 and 1, and setting a decision threshold would yield a concrete prediction. Independent of a decision threshold, the area under the receiver operating characteristic curve (AUC) is a popular metric to evaluate the performance of binary classifiers. This value is between 0 and 1, and the performance of a classification method is better the larger this value is.    

The computational efficiency of the k-NN search algorithm could be significantly enhanced \cite{efficientknn}. Nevertheless, to maintain methodological clarity and establish a baseline for performance evaluation, we employ the standard implementation in our experimental framework.

\section{Experiments}\label{sec:exps}

In the following, we demonstrate the performance of our algorithm on datasets publicly available via Kaggle competitions \cite{kaggle-titanic,kaggle-rossmann}. We tested our framework on the first dataset (less than 1000 rows) via a classification task and on the second (over 1,000,000 rows) via a regression task. For both experiments, we used the Kaggle training set and reserved some rows to serve as a test set in order to guarantee access to target test values and enable to measure the performance. 

We apply Algorithm \ref{alg:embedding_framework}, where we train a one-layer GAE with loss function according to Equation \ref{eq:loss} with $\alpha=0$. We used the GAE python implementation of {\tt torch\_geometric.nn} with a learning rate of 0.01. We restricted the maximal gradient norm to 1e-6 in order to avoid large spikes in the loss function. As stop criterion we used the relative improvement of the loss function after every 10,000 epochs and stopped as soon as this relative improvement is less than 0.1\%.
These parameters were chosen to demonstrate the applicability of our method, they can be adapted and optimized for each dataset.

\subsection{Titanic Dataset}

\begin{figure*}
\includegraphics[width=\textwidth]{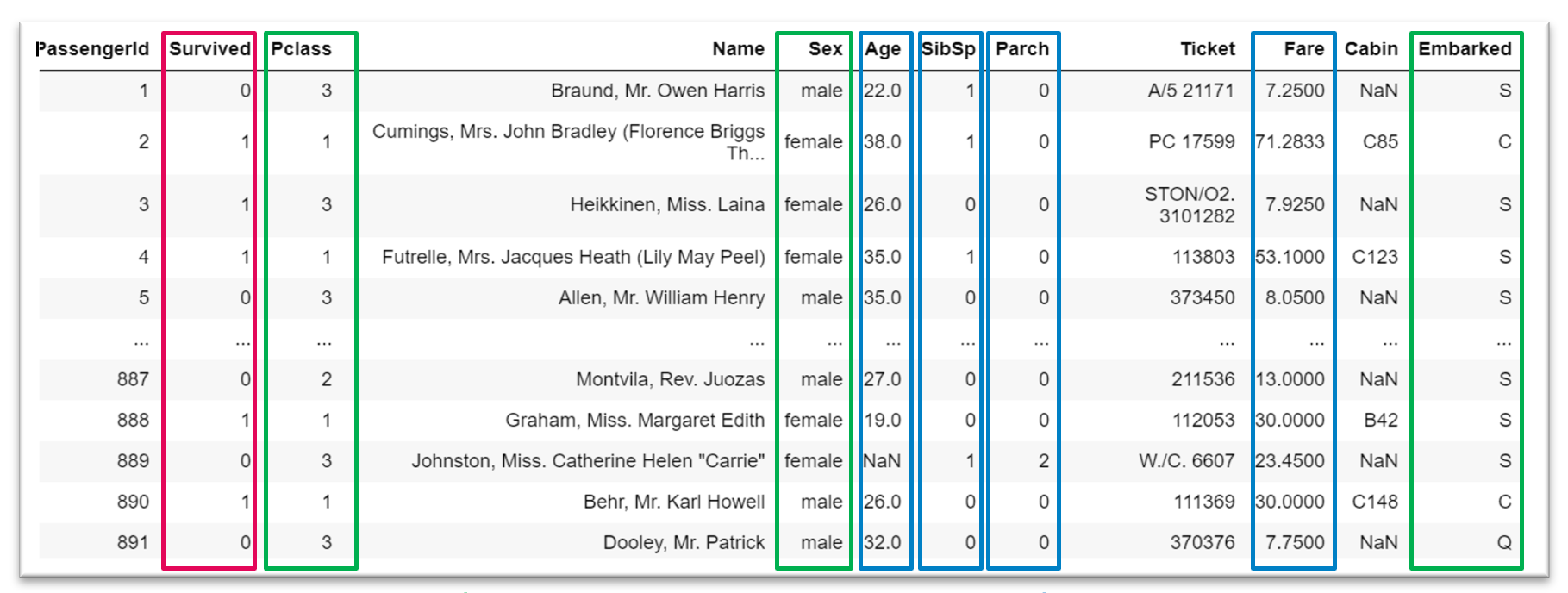}
\caption{Titanic dataset: Columns used for embedding construction are indicated in green (categorical features) and 
blue (numerical features), the target column is marked red.\label{fig:titanic}}
\end{figure*}

Figure~\ref{fig:titanic} shows the data table from the Kaggle competition~\cite{kaggle-titanic}. The goal of this competition is to predict whether a passenger of the sunken Titanic has survived, i.e.\ the target column is \emph{Survived}. We train embeddings, incorporating the categorical columns \emph{Pclass}, \emph{Sex} and \emph{Embarked} and the numerical columns \emph{Age}, \emph{SibSp}, \emph{Parch} and \emph{Fare}. We excluded the columns \emph{PassengerID} and \emph{Name}, since they consist of unique entries and thus do not provide further information regarding the similarity of the rows. Furthermore, we excluded the column \emph{Cabin}, since in the vast majority of rows it is not filled. We randomly selected approximately 10\% of the rows (74) as test samples, the remaining 90\% (817) served as the input table for our embedding algorithm. We divided the values of the numerical columns \emph{Age} and \emph{Fare} into 20 bins, whereas for the other two numerical columns \emph{SibSp} and \emph{Parch} we used one bin for each distinct value.

\begin{figure}
\includegraphics[width=\columnwidth]{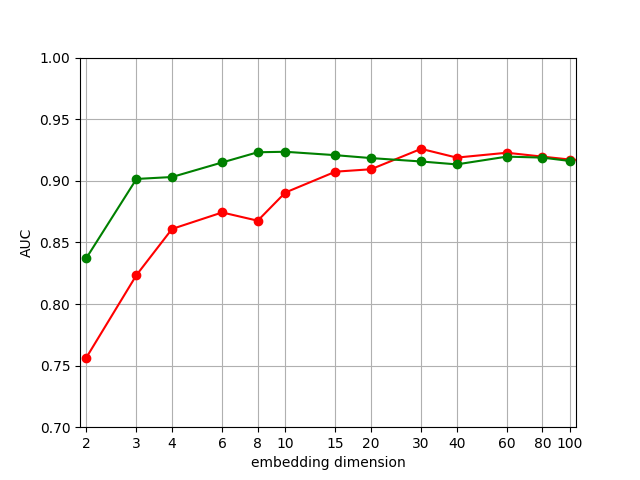}
\includegraphics[width=\columnwidth]{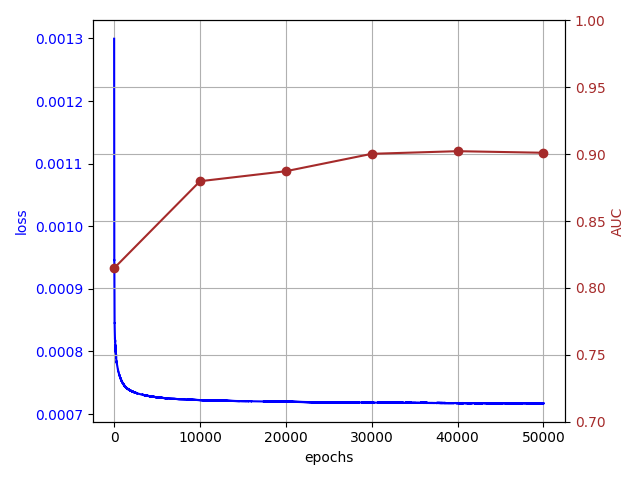}
\caption{Top: Results for the Titanic dataset for our algorithm (green) and for EmbDI-related embeddings (red)
for comparison. Bottom: Progress of the loss function~(\ref{eq:loss}) over the number of epochs (blue) and 
corresponding AUC metric (brown) for embedding dimension $d=3$.\label{fig:titanic_results}} 
\end{figure}

We emphasize that these embeddings are target and task independent and could be used for outlier detection, regression or alternative tasks. To assess the quality, we align with the Kaggle competition. For the classification task of predicting \emph{Survived} we chose the AUC as a quality measure (see Section~\ref{subsec:quality}). The results with respect to different embedding dimensions $d$ are plotted in Figure~\ref{fig:titanic_results} (top). For comparison, we additionally constructed a word2vec embedding of the entities based on 1,000,000 sentences built on the graph in analogy to the EmbDI algorithm~\cite{embdi2020}. Note that the horizontal axis in Figure~\ref{fig:titanic_results} (top) is logarithmically scaled to accommodate the wide range of embedding dimensions. Furthermore, Figure~\ref{fig:titanic_results} (bottom) shows the progress of the loss function~(\ref{eq:loss}) over the number of epochs and the progress of the corresponding AUC metric for comparison.

\subsection{Rossmann Dataset}

\begin{figure}
\includegraphics[width=\columnwidth]{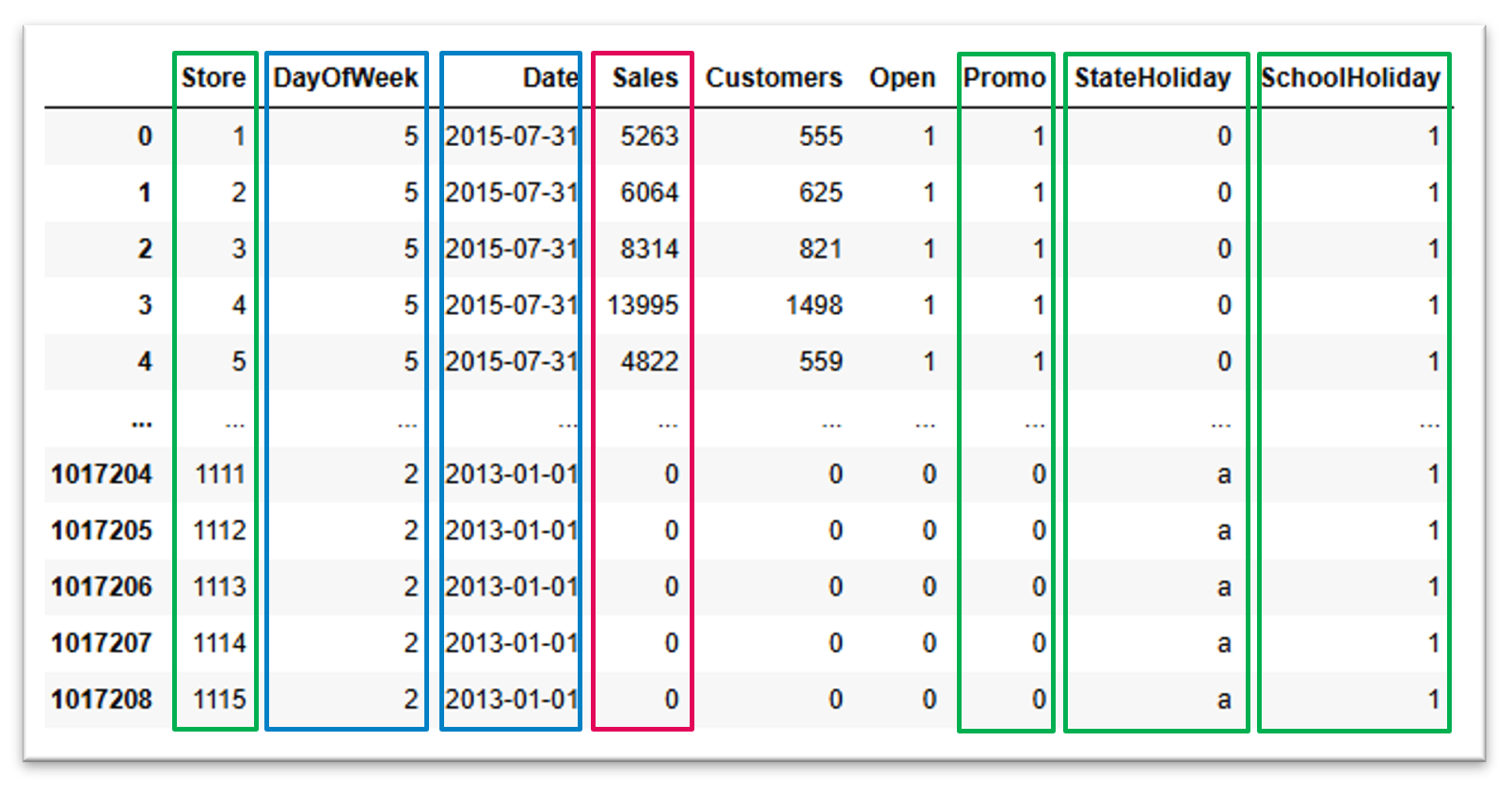}
\caption{Rossmann dataset: Columns used for embedding construction are indicated in green (categorical features) and 
blue (numerical features), the target column is marked red.\label{fig:rossmann}}
\end{figure}

Figure~\ref{fig:rossmann} illustrates data from the Kaggle competition~\cite{kaggle-rossmann}. Here, the target column is \emph{Sales}. For embedding construction, we took into account the categorical columns \emph{Store}, \emph{Promo}, \emph{StateHoliday} and \emph{SchoolHoliday} as well as the numerical columns \emph{DayOfWeek} and \emph{Date}. We excluded closed stores, i.e., rows with \emph{Open}=0, and the column \emph{Customers}, as this information is usually not available a priori to determine the sales. From the remaining 844,392 rows, which are ordered chronologically, we selected the top, i.e. latest approximately 10\% (84,475) as our test set, and the remaining 90\% (759,917) to compute embeddings. The \emph{Date} values (without the year) were assigned to 50 bins, whereas for the column \emph{DayOfWeek} we used one bin for each distinct value.

\begin{figure}[htb]
\includegraphics[width=\columnwidth]{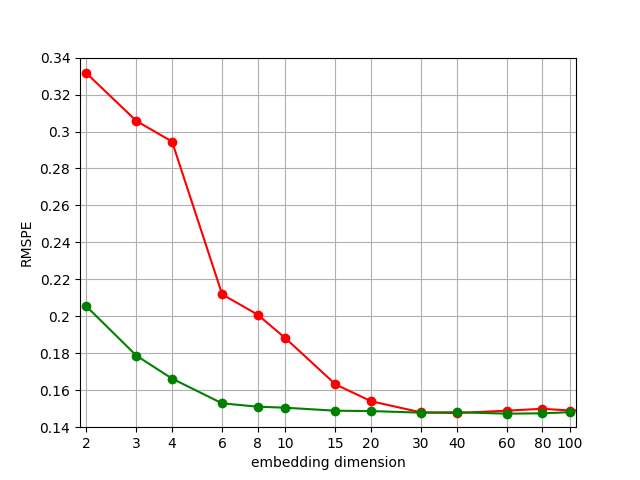}
\includegraphics[width=\columnwidth]{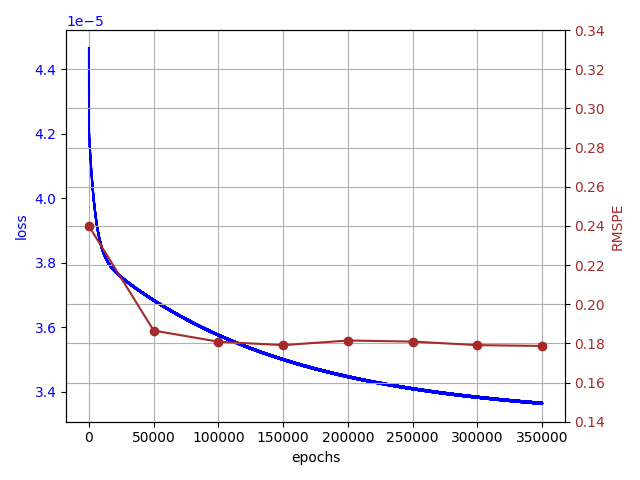}
\caption{Top: Results for the Rossmann dataset for our algorithm (green) and for EmbDI-related embeddings (red)
for comparison. Bottom: Progress of the loss function~(\ref{eq:loss}) over the number of epochs (blue) and 
corresponding RMSPE metric (brown) for embedding dimension $d=3$.\label{fig:rossmann_results}}
\end{figure}

Again, these embeddings are target and task independent and could be used for any application task. To assess the quality, we evaluated the regression task of predicting \emph{Sales} with the RMSPE metric (see Section~\ref{subsec:quality}), aligning with the original Kaggle competition. Note that we did not compare our algorithm against proposed competition methods as they are optimized specifically for this dataset, contradicting with our goal of creating universal embeddings. Since the sales of different stores are not directly comparable, we normalized the \emph{Sales} values by dividing them by the mean value per store, averaged over the training set. Such a normalization does not influence the RMSPE metric, which is a relative metric. The {RMSPE} results with respect to the embedding dimension is plotted in Figure~\ref{fig:rossmann_results} (top), again using a logarithmically scaled horizontal axis. As before, for comparison we calculated the word2vec embedding of the entities based on 1,000,000 sentences according to the EmbDI algorithm~\cite{embdi2020}. Furthermore, Figure~\ref{fig:rossmann_results} (bottom) shows the progress of the loss function~(\ref{eq:loss}) over the number of epochs and the progress of the corresponding RMSPE metric for comparison. 

\section{Discussion}\label{sec:disc}

We benchmark against EmbDI~\cite{embdi2020} because, to the best of our knowledge, it is the only publicly documented method that explicitly aims at producing task-independent row embeddings for arbitrary tables.
Competing baselines such as XGBoost \cite{XGBoost}, LightGBM \cite{LightGBM} or TabNet \cite{Tabnet2021} expose their power only after they have been trained (or re-trained) on a specific supervised objective. Due to the task-specific training, they demonstrate partially superior downstream performance, but do not output reusable embeddings. For instance, XGBoost achieves an AUC of 0.931 for the Titanic classification task and an RMSPE of 0.2153 for the Rossmann regression task. Notably, TabPFN \cite{hollmann2025tabpfn} attains an even higher AUC of 0.946 for the Titanic dataset. Its application is limited to small datasets \cite{hollmann2025tabpfn} and thus it is not applicable to the Rossmann dataset. Although our method provides universal embeddings and is not trained on specific tasks, our results are in the same range.

Figure~\ref{fig:titanic_results} (top) and Figure~\ref{fig:rossmann_results} (top) show for large embedding dimensions that our algorithm has a performance comparable to the EmbDI algorithm. A significant difference is visible for low-dimensional embeddings, where our approach clearly outperforms EmbDI. Moreover, we need fewer dimensions to reach the same performance as for larger dimensions, e.g., for the Rossmann dataset, $d=15$ is sufficient to store the relevant information, whereas EmbDI requires $d=30$ for similar results. This behavior is crucial for industrial applications: When computing embeddings for large-scale datasets to be used for different downstream tasks, these embeddings have to be stored in a vector database. The amount of storage needed for this database is proportional to the embedding dimension. Hence the embedding dimension is desired to be chosen as small as possible. Moreover, the computational effort for constructing low-dimensional embeddings is significantly smaller.

Figure~\ref{fig:titanic_results} (bottom) and Figure~\ref{fig:rossmann_results} (bottom) illustrate the progress of the loss function with respect to the number of epochs. The observation of the decay provides information regarding the choice of the parameter for restricting the maximal gradient norm. When the gradient norm is not restricted or the maximal allowed gradient norm is too large, the loss function exhibits large spikes, leading to a non-optimal convergence. By restricting the maximal gradient norm to a small value (see Section~\ref{sec:exps}), these spikes could be avoided.

Due to the universality of our embeddings, the training loss is independent of the downstream task metric. To ensure that, when the training loss decreases, the downstream task metric also improves, we additionally plotted the downstream task metric with respect to the number of epochs. In both dataset cases, we observe that, in general, further training leads to a better performance. Additionally, even though the training loss is not converged, the task metric seems to reach a plateau which cannot be further improved. Hence the specific definition of the stopping criterion for GAE training seems to be of minor importance.

In both example applications, we considered a single table. An important extension is the combination of information from different tables. In relational databases, a multitude of tables is connected to each other, for instance by foreign keys. When embedding several connected tables, this connecting information must be taken into account. This can be done in different ways: As in the EmbDI algorithm~\cite{embdi2020}, a common graph containing all the information from a multitude of tables can be built. Another possibility would be to embed each table separately and combine the embeddings afterwards. From an industrial point of view, the second approach would be favorable as information, accessible to a later time, could be taken into account. Thus, the embeddings could be improved successively.

\textbf{Limitations and future work:}
We assessed our method using both a small and a larger dataset, applying it to classification and regression. Future research should incorporate a broader range of datasets to enable a more comprehensive performance comparison.
Expanding the method to accommodate modalities such as text, images, or time series data is also a valuable direction for future work.  Furthermore, as mentioned in the previous paragraph, the applicability of our method to relational tables should be further investigated.

\section{Conclusion}\label{sec:concl}

In this paper, we highlighted the problem of developing tabular processing tools in an industrial setting where tasks and targets are not defined beforehand. To address this, we presented an algorithm for constructing universal embeddings of tabular data, which can be applied in industrial environments storing data in large relational databases. Our approach yields target- and task-independent embeddings that, subsequently, can be applied in a variety of downstream tasks such as classification, regression, similarity search or outlier detection. Our method is based on an autoencoder strategy for graphs, representing the tabular data. We decided on a two-step approach: First, we construct entity embeddings. Second, for each table row, i.e.\ each sample, we aggregate the corresponding entity embeddings. The entity embeddings are reusable for similar unseen data, allowing for continuous applicability without new training. The row embeddings are applied for arbitrary downstream tasks exploiting their distance in the vector space. For an exemplary classification and regression task, we demonstrated that the necessary embedding dimension for our algorithm is significantly smaller compared to state-of-the-art methods. This is an important advantage for industrial applications with very large data tables. Our approach allows for various techniques to merge the information of two or more connected tables. However, further research on how to combine the table information and the corresponding evaluation is necessary.

\clearpage

\bibliographystyle{ACM-Reference-Format}
\bibliography{sample}


\begin{thebibliography}{35}

%%% the Web address.

\ifx \showCODEN    \undefined \def \showCODEN     #1{\unskip}     \fi
\ifx \showDOI      \undefined \def \showDOI       #1{#1}\fi
\ifx \showISBNx    \undefined \def \showISBNx     #1{\unskip}     \fi
\ifx \showISBNxiii \undefined \def \showISBNxiii  #1{\unskip}     \fi
\ifx \showISSN     \undefined \def \showISSN      #1{\unskip}     \fi
\ifx \showLCCN     \undefined \def \showLCCN      #1{\unskip}     \fi
\ifx \shownote     \undefined \def \shownote      #1{#1}          \fi
\ifx \showarticletitle \undefined \def \showarticletitle #1{#1}   \fi
\ifx \showURL      \undefined \def \showURL       {\relax}        \fi
% invisible to TeX
\providecommand\bibfield[2]{#2}
\providecommand\bibinfo[2]{#2}
\providecommand\natexlab[1]{#1}
\providecommand\showeprint[2][]{arXiv:#2}

\bibitem[\protect\citeauthoryear{Arik and Pfister}{Arik and Pfister}{2021}]%
        {Tabnet2021}
\bibfield{author}{\bibinfo{person}{Sercan~\"O. Arik} {and}
  \bibinfo{person}{Tomas Pfister}.} \bibinfo{year}{2021}\natexlab{}.
\newblock \showarticletitle{{TabNet}: Attentive Interpretable Tabular
  Learning}.
\newblock \bibinfo{journal}{\emph{Proceedings of the AAAI Conference on
  Artificial Intelligence}} \bibinfo{volume}{35}, \bibinfo{number}{8}
  (\bibinfo{year}{2021}), \bibinfo{pages}{6679--6687}.
\newblock
\urldef\tempurl%
\url{https://doi.org/10.1609/aaai.v35i8.16826}
\showDOI{\tempurl}


\bibitem[\protect\citeauthoryear{Borisov, Leemann, Seßler, Haug, Pawelczyk,
  and Kasneci}{Borisov et~al\mbox{.}}{2024}]%
        {surveyBorisov}
\bibfield{author}{\bibinfo{person}{Vadim Borisov}, \bibinfo{person}{Tobias
  Leemann}, \bibinfo{person}{Kathrin Seßler}, \bibinfo{person}{Johannes Haug},
  \bibinfo{person}{Martin Pawelczyk}, {and} \bibinfo{person}{Gjergji Kasneci}.}
  \bibinfo{year}{2024}\natexlab{}.
\newblock \showarticletitle{Deep Neural Networks and Tabular Data: A Survey}.
\newblock \bibinfo{journal}{\emph{IEEE Transactions on Neural Networks and
  Learning Systems}} \bibinfo{volume}{35}, \bibinfo{number}{6}
  (\bibinfo{year}{2024}), \bibinfo{pages}{7499--7519}.
\newblock
\urldef\tempurl%
\url{https://doi.org/10.1109/TNNLS.2022.3229161}
\showDOI{\tempurl}


\bibitem[\protect\citeauthoryear{Cappuzzo, Papotti, and
  Thirumuruganathan}{Cappuzzo et~al\mbox{.}}{2020}]%
        {embdi2020}
\bibfield{author}{\bibinfo{person}{Riccardo Cappuzzo}, \bibinfo{person}{Paolo
  Papotti}, {and} \bibinfo{person}{Saravanan Thirumuruganathan}.}
  \bibinfo{year}{2020}\natexlab{}.
\newblock \showarticletitle{Creating Embeddings of Heterogeneous Relational
  Datasets for Data Integration Tasks}. In \bibinfo{booktitle}{\emph{ACM
  SIGMOD/PODS Conference}}. \bibinfo{pages}{1335--1349}.
\newblock
\urldef\tempurl%
\url{https://doi.org/10.1145/3318464.3389742}
\showDOI{\tempurl}


\bibitem[\protect\citeauthoryear{Chen and Guestrin}{Chen and Guestrin}{2016}]%
        {XGBoost}
\bibfield{author}{\bibinfo{person}{Tianqi Chen} {and} \bibinfo{person}{Carlos
  Guestrin}.} \bibinfo{year}{2016}\natexlab{}.
\newblock \showarticletitle{{XGBoost}: A Scalable Tree Boosting System}. In
  \bibinfo{booktitle}{\emph{Proceedings of the 22nd ACM SIGKDD International
  Conference on Knowledge Discovery and Data Mining}} (San Francisco,
  California, USA) \emph{(\bibinfo{series}{KDD '16})}.
  \bibinfo{publisher}{Association for Computing Machinery},
  \bibinfo{address}{New York, NY, USA}, \bibinfo{pages}{785–794}.
\newblock
\showISBNx{9781450342322}
\urldef\tempurl%
\url{https://doi.org/10.1145/2939672.2939785}
\showDOI{\tempurl}


\bibitem[\protect\citeauthoryear{Chernov}{Chernov}{2025}]%
        {chernov2025ggmoevsmlp}
\bibfield{author}{\bibinfo{person}{Andrei Chernov}.}
  \bibinfo{year}{2025}\natexlab{}.
\newblock \bibinfo{title}{{(GG) MoE vs. MLP on Tabular Data}}.
\newblock
\newblock
\urldef\tempurl%
\url{https://arxiv.org/abs/2502.03608}
\showURL{%
\tempurl}


\bibitem[\protect\citeauthoryear{Cong, Hulsebos, Sun, Groth, and Jagadish}{Cong
  et~al\mbox{.}}{2024}]%
        {OBSERVATORY}
\bibfield{author}{\bibinfo{person}{Tianji Cong}, \bibinfo{person}{Madelon
  Hulsebos}, \bibinfo{person}{Zhenjie Sun}, \bibinfo{person}{Paul Groth}, {and}
  \bibinfo{person}{H.~V. Jagadish}.} \bibinfo{year}{2024}\natexlab{}.
\newblock \showarticletitle{Observatory: Characterizing Embeddings of
  Relational Tables}.
\newblock \bibinfo{journal}{\emph{Proc. VLDB Endow.}} \bibinfo{volume}{17},
  \bibinfo{number}{4} (\bibinfo{year}{2024}), \bibinfo{pages}{849--862}.
\newblock
\showISSN{2150-8097}
\urldef\tempurl%
\url{https://doi.org/10.14778/3636218.3636237}
\showDOI{\tempurl}


\bibitem[\protect\citeauthoryear{Cukierski}{Cukierski}{2012}]%
        {kaggle-titanic}
\bibfield{author}{\bibinfo{person}{Will Cukierski}.}
  \bibinfo{year}{2012}\natexlab{}.
\newblock \bibinfo{title}{Titanic - Machine Learning from Disaster}.
\newblock
\newblock
\urldef\tempurl%
\url{https://www.kaggle.com/competitions/titanic}
\showURL{%
\tempurl}


\bibitem[\protect\citeauthoryear{Deng, Sun, Lees, Wu, and Yu}{Deng
  et~al\mbox{.}}{2020}]%
        {TURL}
\bibfield{author}{\bibinfo{person}{Xiang Deng}, \bibinfo{person}{Huan Sun},
  \bibinfo{person}{Alyssa Lees}, \bibinfo{person}{You Wu}, {and}
  \bibinfo{person}{Cong Yu}.} \bibinfo{year}{2020}\natexlab{}.
\newblock \showarticletitle{{TURL}: table understanding through representation
  learning}.
\newblock \bibinfo{journal}{\emph{Proc. VLDB Endow.}} \bibinfo{volume}{14},
  \bibinfo{number}{3} (\bibinfo{date}{Nov.} \bibinfo{year}{2020}),
  \bibinfo{pages}{307–319}.
\newblock
\showISSN{2150-8097}
\urldef\tempurl%
\url{https://doi.org/10.14778/3430915.3430921}
\showDOI{\tempurl}


\bibitem[\protect\citeauthoryear{Gorishniy, Kotelnikov, and Babenko}{Gorishniy
  et~al\mbox{.}}{2025}]%
        {gorishniy2025tabm}
\bibfield{author}{\bibinfo{person}{Yury Gorishniy}, \bibinfo{person}{Akim
  Kotelnikov}, {and} \bibinfo{person}{Artem Babenko}.}
  \bibinfo{year}{2025}\natexlab{}.
\newblock \showarticletitle{{TabM}: Advancing tabular deep learning with
  parameter-efficient ensembling}. In \bibinfo{booktitle}{\emph{The Thirteenth
  International Conference on Learning Representations}}.
\newblock
\urldef\tempurl%
\url{https://openreview.net/forum?id=Sd4wYYOhmY}
\showURL{%
\tempurl}


\bibitem[\protect\citeauthoryear{Grinsztajn, Oyallon, and Varoquaux}{Grinsztajn
  et~al\mbox{.}}{2022}]%
        {whyTreeBetterDL}
\bibfield{author}{\bibinfo{person}{L\'{e}o Grinsztajn},
  \bibinfo{person}{Edouard Oyallon}, {and} \bibinfo{person}{Ga\"{e}l
  Varoquaux}.} \bibinfo{year}{2022}\natexlab{}.
\newblock \showarticletitle{Why do tree-based models still outperform deep
  learning on typical tabular data?}. In \bibinfo{booktitle}{\emph{Proceedings
  of the 36th International Conference on Neural Information Processing
  Systems}} (New Orleans, LA, USA) \emph{(\bibinfo{series}{NIPS '22})}.
  \bibinfo{publisher}{Curran Associates Inc.}, \bibinfo{address}{Red Hook, NY,
  USA}, Article \bibinfo{articleno}{37}, \bibinfo{numpages}{14}~pages.
\newblock
\showISBNx{9781713871088}


\bibitem[\protect\citeauthoryear{Hollmann, M{\"u}ller, Purucker, Krishnakumar,
  K{\"o}rfer, Hoo, Schirrmeister, and Hutter}{Hollmann et~al\mbox{.}}{2025}]%
        {hollmann2025tabpfn}
\bibfield{author}{\bibinfo{person}{Noah Hollmann}, \bibinfo{person}{Samuel
  M{\"u}ller}, \bibinfo{person}{Lennart Purucker}, \bibinfo{person}{Arjun
  Krishnakumar}, \bibinfo{person}{Max K{\"o}rfer}, \bibinfo{person}{Shi~Bin
  Hoo}, \bibinfo{person}{Robin~Tibor Schirrmeister}, {and}
  \bibinfo{person}{Frank Hutter}.} \bibinfo{year}{2025}\natexlab{}.
\newblock \showarticletitle{Accurate predictions on small data with a tabular
  foundation model}.
\newblock \bibinfo{journal}{\emph{Nature}} (\bibinfo{date}{09 01}
  \bibinfo{year}{2025}).
\newblock
\urldef\tempurl%
\url{https://doi.org/10.1038/s41586-024-08328-6}
\showDOI{\tempurl}


\bibitem[\protect\citeauthoryear{Huang, Khetan, Cvitkovic, and Karnin}{Huang
  et~al\mbox{.}}{2020}]%
        {Tabtransformer}
\bibfield{author}{\bibinfo{person}{Xin Huang}, \bibinfo{person}{Ashish Khetan},
  \bibinfo{person}{Milan Cvitkovic}, {and} \bibinfo{person}{Zohar Karnin}.}
  \bibinfo{year}{2020}\natexlab{}.
\newblock \bibinfo{title}{{TabTransformer}: Tabular Data Modeling Using
  Contextual Embeddings}.
\newblock
\newblock
\urldef\tempurl%
\url{https://doi.org/10.48550/arXiv.2012.06678}
\showDOI{\tempurl}


\bibitem[\protect\citeauthoryear{Jiang, Liu, Cai, Zhou, and Ye}{Jiang
  et~al\mbox{.}}{2025a}]%
        {jiang2025representationlearningtabulardata}
\bibfield{author}{\bibinfo{person}{Jun-Peng Jiang}, \bibinfo{person}{Si-Yang
  Liu}, \bibinfo{person}{Hao-Run Cai}, \bibinfo{person}{Qile Zhou}, {and}
  \bibinfo{person}{Han-Jia Ye}.} \bibinfo{year}{2025}\natexlab{a}.
\newblock \bibinfo{title}{Representation Learning for Tabular Data: A
  Comprehensive Survey}.
\newblock
\newblock
\urldef\tempurl%
\url{https://arxiv.org/abs/2504.16109}
\showURL{%
\tempurl}


\bibitem[\protect\citeauthoryear{Jiang, Liu, Cai, Zhou, and Ye}{Jiang
  et~al\mbox{.}}{2025b}]%
        {TARTE}
\bibfield{author}{\bibinfo{person}{Jun-Peng Jiang}, \bibinfo{person}{Si-Yang
  Liu}, \bibinfo{person}{Hao-Run Cai}, \bibinfo{person}{Qile Zhou}, {and}
  \bibinfo{person}{Han-Jia Ye}.} \bibinfo{year}{2025}\natexlab{b}.
\newblock \bibinfo{title}{Table Foundation Models: on knowledge pre-training
  for tabular learning}.
\newblock
\newblock
\urldef\tempurl%
\url{https://arxiv.org/abs/2505.14415v1}
\showURL{%
\tempurl}


\bibitem[\protect\citeauthoryear{Ke, Meng, Finley, Wang, Chen, Ma, Ye, and
  Liu}{Ke et~al\mbox{.}}{2017}]%
        {LightGBM}
\bibfield{author}{\bibinfo{person}{Guolin Ke}, \bibinfo{person}{Qi Meng},
  \bibinfo{person}{Thomas Finley}, \bibinfo{person}{Taifeng Wang},
  \bibinfo{person}{Wei Chen}, \bibinfo{person}{Weidong Ma},
  \bibinfo{person}{Qiwei Ye}, {and} \bibinfo{person}{Tie-Yan Liu}.}
  \bibinfo{year}{2017}\natexlab{}.
\newblock \showarticletitle{{LightGBM}: a highly efficient gradient boosting
  decision tree}. In \bibinfo{booktitle}{\emph{Proceedings of the 31st
  International Conference on Neural Information Processing Systems}} (Long
  Beach, California, USA) \emph{(\bibinfo{series}{NIPS'17})}.
  \bibinfo{publisher}{Curran Associates Inc.}, \bibinfo{address}{Red Hook, NY,
  USA}, \bibinfo{pages}{3149–3157}.
\newblock
\showISBNx{9781510860964}


\bibitem[\protect\citeauthoryear{Kim, Grinsztajn, and Varoquaux}{Kim
  et~al\mbox{.}}{2024}]%
        {CARTE}
\bibfield{author}{\bibinfo{person}{Myung~Jun Kim}, \bibinfo{person}{L\'eo
  Grinsztajn}, {and} \bibinfo{person}{Ga{\"e}l Varoquaux}.}
  \bibinfo{year}{2024}\natexlab{}.
\newblock \showarticletitle{{CARTE}: Pretraining and Transfer for Tabular
  Learning}. In \bibinfo{booktitle}{\emph{Proceedings of the 41st International
  Conference on Machine Learning}} (Vienna, Austria),
  Vol.~\bibinfo{volume}{PMLR 235}. \bibinfo{pages}{23843--23866}.
\newblock


\bibitem[\protect\citeauthoryear{Kipf and Welling}{Kipf and Welling}{2016}]%
        {kipf2016variationalgraphautoencoders}
\bibfield{author}{\bibinfo{person}{Thomas~N. Kipf} {and} \bibinfo{person}{Max
  Welling}.} \bibinfo{year}{2016}\natexlab{}.
\newblock \bibinfo{title}{{Variational Graph Auto-Encoder}s}.
\newblock
\newblock
\urldef\tempurl%
\url{https://arxiv.org/abs/1611.07308}
\showURL{%
\tempurl}


\bibitem[\protect\citeauthoryear{Kipf and Welling}{Kipf and Welling}{2017}]%
        {kipf2017semisupervised}
\bibfield{author}{\bibinfo{person}{Thomas~N. Kipf} {and} \bibinfo{person}{Max
  Welling}.} \bibinfo{year}{2017}\natexlab{}.
\newblock \showarticletitle{Semi-Supervised Classification with Graph
  Convolutional Networks}. In \bibinfo{booktitle}{\emph{International
  Conference on Learning Representations}}.
\newblock
\urldef\tempurl%
\url{https://openreview.net/forum?id=SJU4ayYgl}
\showURL{%
\tempurl}


\bibitem[\protect\citeauthoryear{Knauer and Cukierski}{Knauer and
  Cukierski}{2015}]%
        {kaggle-rossmann}
\bibfield{author}{\bibinfo{person}{Florian Knauer} {and} \bibinfo{person}{Will
  Cukierski}.} \bibinfo{year}{2015}\natexlab{}.
\newblock \bibinfo{title}{Rossmann Store Sales}.
\newblock
\newblock
\urldef\tempurl%
\url{https://kaggle.com/competitions/rossmann-store-sales}
\showURL{%
\tempurl}


\bibitem[\protect\citeauthoryear{Malkov and Yashunin}{Malkov and
  Yashunin}{2020}]%
        {efficientknn}
\bibfield{author}{\bibinfo{person}{Yu~A. Malkov} {and} \bibinfo{person}{D.~A.
  Yashunin}.} \bibinfo{year}{2020}\natexlab{}.
\newblock \showarticletitle{Efficient and Robust Approximate Nearest Neighbor
  Search Using Hierarchical Navigable Small World Graphs}.
\newblock \bibinfo{journal}{\emph{IEEE Transactions on Pattern Analysis and
  Machine Intelligence}} \bibinfo{volume}{42}, \bibinfo{number}{4}
  (\bibinfo{year}{2020}), \bibinfo{pages}{824--836}.
\newblock
\urldef\tempurl%
\url{https://doi.org/10.1109/TPAMI.2018.2889473}
\showDOI{\tempurl}


\bibitem[\protect\citeauthoryear{Mikolov, Chen, Corrado, and Dean}{Mikolov
  et~al\mbox{.}}{2013}]%
        {word2vec}
\bibfield{author}{\bibinfo{person}{Tomas Mikolov}, \bibinfo{person}{Kai Chen},
  \bibinfo{person}{Greg Corrado}, {and} \bibinfo{person}{Jeffrey Dean}.}
  \bibinfo{year}{2013}\natexlab{}.
\newblock \bibinfo{title}{Efficient Estimation of Word Representations in
  Vector Space}.
\newblock
\newblock
\urldef\tempurl%
\url{https://arxiv.org/abs/1301.3781}
\showURL{%
\tempurl}


\bibitem[\protect\citeauthoryear{Pennington, Socher, and Manning}{Pennington
  et~al\mbox{.}}{2014}]%
        {glove}
\bibfield{author}{\bibinfo{person}{Jeffrey Pennington},
  \bibinfo{person}{Richard Socher}, {and} \bibinfo{person}{Christopher
  Manning}.} \bibinfo{year}{2014}\natexlab{}.
\newblock \showarticletitle{Glove: Global Vectors for Word Representation}.
\newblock \bibinfo{journal}{\emph{EMNLP}}  \bibinfo{volume}{14}
  (\bibinfo{year}{2014}), \bibinfo{pages}{1532--1543}.
\newblock
\urldef\tempurl%
\url{https://doi.org/10.3115/v1/D14-1162}
\showDOI{\tempurl}


\bibitem[\protect\citeauthoryear{Prokhorenkova, Gusev, Vorobev, Dorogush, and
  Gulin}{Prokhorenkova et~al\mbox{.}}{2018}]%
        {CatBoost}
\bibfield{author}{\bibinfo{person}{Liudmila Prokhorenkova},
  \bibinfo{person}{Gleb Gusev}, \bibinfo{person}{Aleksandr Vorobev},
  \bibinfo{person}{Anna~Veronika Dorogush}, {and} \bibinfo{person}{Andrey
  Gulin}.} \bibinfo{year}{2018}\natexlab{}.
\newblock \showarticletitle{{CatBoost}: unbiased boosting with categorical
  features}. In \bibinfo{booktitle}{\emph{Proceedings of the 32nd International
  Conference on Neural Information Processing Systems}} (Montr\'{e}al, Canada)
  \emph{(\bibinfo{series}{NIPS'18})}. \bibinfo{publisher}{Curran Associates
  Inc.}, \bibinfo{address}{Red Hook, NY, USA}, \bibinfo{pages}{6639–6649}.
\newblock


\bibitem[\protect\citeauthoryear{Qu, David~Holzm{\"u}ller, and Morvan}{Qu
  et~al\mbox{.}}{2025}]%
        {TabICL}
\bibfield{author}{\bibinfo{person}{Jingang Qu},
  \bibinfo{person}{Ga{\"e}l~Varoquaux David~Holzm{\"u}ller}, {and}
  \bibinfo{person}{Marine~Le Morvan}.} \bibinfo{year}{2025}\natexlab{}.
\newblock \bibinfo{title}{TabICL: A Tabular Foundation Model for In-Context
  Learning on Large Data}.
\newblock
\newblock
\urldef\tempurl%
\url{https://arxiv.org/abs/2502.05564}
\showURL{%
\tempurl}


\bibitem[\protect\citeauthoryear{Salehi and Davulcu}{Salehi and
  Davulcu}{2020}]%
        {GATE}
\bibfield{author}{\bibinfo{person}{Amin Salehi} {and} \bibinfo{person}{Hasan
  Davulcu}.} \bibinfo{year}{2020}\natexlab{}.
\newblock \showarticletitle{Graph Attention Auto-Encoders}. In
  \bibinfo{booktitle}{\emph{2020 IEEE 32nd International Conference on Tools
  with Artificial Intelligence (ICTAI)}}. \bibinfo{pages}{989--996}.
\newblock
\urldef\tempurl%
\url{https://doi.org/10.1109/ICTAI50040.2020.00154}
\showDOI{\tempurl}


\bibitem[\protect\citeauthoryear{Shwartz-Ziv and Armon}{Shwartz-Ziv and
  Armon}{2022}]%
        {DLnotallyouneed}
\bibfield{author}{\bibinfo{person}{Ravid Shwartz-Ziv} {and}
  \bibinfo{person}{Amitai Armon}.} \bibinfo{year}{2022}\natexlab{}.
\newblock \showarticletitle{Tabular data: Deep learning is not all you need}.
\newblock \bibinfo{journal}{\emph{Inf. Fusion}} \bibinfo{volume}{81},
  \bibinfo{number}{C} (\bibinfo{year}{2022}), \bibinfo{pages}{84–90}.
\newblock
\showISSN{1566-2535}
\urldef\tempurl%
\url{https://doi.org/10.1016/j.inffus.2021.11.011}
\showDOI{\tempurl}


\bibitem[\protect\citeauthoryear{Singh and Bedathur}{Singh and
  Bedathur}{2023}]%
        {singh2023embeddingstabulardatasurvey}
\bibfield{author}{\bibinfo{person}{Rajat Singh} {and} \bibinfo{person}{Srikanta
  Bedathur}.} \bibinfo{year}{2023}\natexlab{}.
\newblock \bibinfo{title}{Embeddings for Tabular Data: A Survey}.
\newblock
\newblock
\urldef\tempurl%
\url{https://arxiv.org/abs/2302.11777}
\showURL{%
\tempurl}


\bibitem[\protect\citeauthoryear{Somepalli, Goldblum, Schwarzschild, Bruss, and
  Goldstein}{Somepalli et~al\mbox{.}}{2021}]%
        {saint}
\bibfield{author}{\bibinfo{person}{Gowthami Somepalli}, \bibinfo{person}{Micah
  Goldblum}, \bibinfo{person}{Avi Schwarzschild}, \bibinfo{person}{C. Bruss},
  {and} \bibinfo{person}{Tom Goldstein}.} \bibinfo{year}{2021}\natexlab{}.
\newblock \bibinfo{title}{{SAINT}: Improved Neural Networks for Tabular Data
  via Row Attention and Contrastive Pre-Training}.
\newblock
\newblock
\urldef\tempurl%
\url{https://doi.org/10.48550/arXiv.2106.01342}
\showDOI{\tempurl}


\bibitem[\protect\citeauthoryear{Tan, Gui, and Qiu}{Tan et~al\mbox{.}}{2024}]%
        {GAEFS}
\bibfield{author}{\bibinfo{person}{Jun Tan}, \bibinfo{person}{Ning Gui}, {and}
  \bibinfo{person}{Zhifeng Qiu}.} \bibinfo{year}{2024}\natexlab{}.
\newblock \showarticletitle{{GAEFS}: Self-supervised Graph Auto-encoder
  enhanced Feature Selection}.
\newblock \bibinfo{journal}{\emph{Knowledge-Based Systems}}
  \bibinfo{volume}{290} (\bibinfo{year}{2024}), \bibinfo{pages}{111523}.
\newblock
\showISSN{0950-7051}
\urldef\tempurl%
\url{https://doi.org/10.1016/j.knosys.2024.111523}
\showDOI{\tempurl}


\bibitem[\protect\citeauthoryear{Vaswani, Shazeer, Parmar, Uszkoreit, Jones,
  Gomez, Kaiser, and Polosukhin}{Vaswani et~al\mbox{.}}{2017}]%
        {Attentionisallyouneed}
\bibfield{author}{\bibinfo{person}{Ashish Vaswani}, \bibinfo{person}{Noam
  Shazeer}, \bibinfo{person}{Niki Parmar}, \bibinfo{person}{Jakob Uszkoreit},
  \bibinfo{person}{Llion Jones}, \bibinfo{person}{Aidan~N. Gomez},
  \bibinfo{person}{\L{}ukasz Kaiser}, {and} \bibinfo{person}{Illia
  Polosukhin}.} \bibinfo{year}{2017}\natexlab{}.
\newblock \showarticletitle{Attention is all you need}. In
  \bibinfo{booktitle}{\emph{Proceedings of the 31st International Conference on
  Neural Information Processing Systems}} (Long Beach, California, USA)
  \emph{(\bibinfo{series}{NIPS'17})}. \bibinfo{publisher}{Curran Associates
  Inc.}, \bibinfo{address}{Red Hook, NY, USA}, \bibinfo{pages}{6000–6010}.
\newblock
\showISBNx{9781510860964}


\bibitem[\protect\citeauthoryear{Villaizán-Vallelado, Salvatori, Carro, and
  Sanchez-Esguevillas}{Villaizán-Vallelado et~al\mbox{.}}{2024}]%
        {VILLAIZANVALLELADO2024106180}
\bibfield{author}{\bibinfo{person}{Mario Villaizán-Vallelado},
  \bibinfo{person}{Matteo Salvatori}, \bibinfo{person}{Belén Carro}, {and}
  \bibinfo{person}{Antonio~Javier Sanchez-Esguevillas}.}
  \bibinfo{year}{2024}\natexlab{}.
\newblock \showarticletitle{Graph Neural Network contextual embedding for Deep
  Learning on tabular data}.
\newblock \bibinfo{journal}{\emph{Neural Networks}}  \bibinfo{volume}{173}
  (\bibinfo{year}{2024}), \bibinfo{pages}{106180}.
\newblock
\showISSN{0893-6080}
\urldef\tempurl%
\url{https://doi.org/10.1016/j.neunet.2024.106180}
\showDOI{\tempurl}


\bibitem[\protect\citeauthoryear{Wen, Tran, and Ba}{Wen et~al\mbox{.}}{2020}]%
        {Wen2020BatchEnsemble}
\bibfield{author}{\bibinfo{person}{Yeming Wen}, \bibinfo{person}{Dustin Tran},
  {and} \bibinfo{person}{Jimmy Ba}.} \bibinfo{year}{2020}\natexlab{}.
\newblock \showarticletitle{{BatchEnsemble}: an Alternative Approach to
  Efficient Ensemble and Lifelong Learning}. In
  \bibinfo{booktitle}{\emph{International Conference on Learning
  Representations}}.
\newblock
\urldef\tempurl%
\url{https://openreview.net/forum?id=Sklf1yrYDr}
\showURL{%
\tempurl}


\bibitem[\protect\citeauthoryear{Wu, Pan, Chen, Long, Zhang, and Yu}{Wu
  et~al\mbox{.}}{2021}]%
        {surveyGNN}
\bibfield{author}{\bibinfo{person}{Zonghan Wu}, \bibinfo{person}{Shirui Pan},
  \bibinfo{person}{Fengwen Chen}, \bibinfo{person}{Guodong Long},
  \bibinfo{person}{Chengqi Zhang}, {and} \bibinfo{person}{Philip~S. Yu}.}
  \bibinfo{year}{2021}\natexlab{}.
\newblock \showarticletitle{A Comprehensive Survey on Graph Neural Networks}.
\newblock \bibinfo{journal}{\emph{IEEE Transactions on Neural Networks and
  Learning Systems}} \bibinfo{volume}{32}, \bibinfo{number}{1}
  (\bibinfo{year}{2021}), \bibinfo{pages}{4--24}.
\newblock
\urldef\tempurl%
\url{https://doi.org/10.1109/TNNLS.2020.2978386}
\showDOI{\tempurl}


\bibitem[\protect\citeauthoryear{Ye, Lu, Wang, Li, Wu, Chen, and Zhao}{Ye
  et~al\mbox{.}}{2024}]%
        {ctBert}
\bibfield{author}{\bibinfo{person}{Chao Ye}, \bibinfo{person}{Guoshan Lu},
  \bibinfo{person}{Haobo Wang}, \bibinfo{person}{Liyao Li},
  \bibinfo{person}{Sai Wu}, \bibinfo{person}{Gang Chen}, {and}
  \bibinfo{person}{Junbo Zhao}.} \bibinfo{year}{2024}\natexlab{}.
\newblock \bibinfo{title}{Towards Cross-Table Masked Pretraining for Web Data
  Mining}.
\newblock
\newblock
\urldef\tempurl%
\url{https://arxiv.org/abs/2307.04308}
\showURL{%
\tempurl}


\bibitem[\protect\citeauthoryear{Yin, Neubig, Yih, and Riedel}{Yin
  et~al\mbox{.}}{2020}]%
        {yin-etal-2020-tabert}
\bibfield{author}{\bibinfo{person}{Pengcheng Yin}, \bibinfo{person}{Graham
  Neubig}, \bibinfo{person}{Wen-tau Yih}, {and} \bibinfo{person}{Sebastian
  Riedel}.} \bibinfo{year}{2020}\natexlab{}.
\newblock \showarticletitle{{T}a{BERT}: Pretraining for Joint Understanding of
  Textual and Tabular Data}. In \bibinfo{booktitle}{\emph{Proceedings of the
  58th Annual Meeting of the Association for Computational Linguistics}}.
  \bibinfo{pages}{8413--8426}.
\newblock
\urldef\tempurl%
\url{https://doi.org/10.18653/v1/2020.acl-main.745}
\showDOI{\tempurl}


\end{thebibliography}

\end{document}